%% file: 00_0main.tex
\documentclass[manuscript]{acmart}
\AtBeginDocument{%
  \providecommand\BibTeX{{%
    \normalfont B\kern-0.5em{\scshape i\kern-0.25em b}\kern-0.8em\TeX}}}

\usepackage{adjustbox}
\usepackage{graphics}
\usepackage{algorithm}
\usepackage{algpseudocode}
\usepackage{gensymb}
\usepackage{breqn}
\usepackage{array, multirow,graphicx}
\usepackage{xcolor}
\usepackage{graphicx} 
\usepackage{caption,subcaption}
\usepackage{natbib}
\usepackage{color}
\usepackage{booktabs}
\usepackage{enumitem}
\usepackage{natbib}
\usepackage{nicefrac}     
\usepackage{microtype}  
\usepackage{lipsum}	
\usepackage{hhline}
\usepackage{subcaption}
\usepackage{graphicx} 
\usepackage{graphicx}
\usepackage{float}
\usepackage{array}
\usepackage{adjustbox}
\usepackage{colortbl}
\usepackage{multirow}
\usepackage{mathtools}
\usepackage{longtable}
\usepackage{threeparttable}
\usepackage{wrapfig}
\usepackage{xspace}
\usepackage{subcaption}
\usepackage{tabularx}
\usepackage{url}
\usepackage{amsmath}

\usepackage{amssymb}
\usepackage{tikz}

\usepackage{algorithm}
\usepackage{algpseudocode}

\usepackage{amsmath}
\DeclareMathOperator*{\argmax}{arg\,max}

\newcommand{\sysname}{\textbf{adaPARL}\xspace}

\usepackage[font=small, skip=0.5ex]{caption}
\usepackage[subtle, tracking=normal]{savetrees}
\copyrightyear{2023}
\acmYear{2023}
\setcopyright{rightsretained}
\acmConference[IoTDI '23]{International Conference on Internet-of-Things Design and Implementation}{May 9--12, 2023}{San Antonio, TX, USA}
\acmBooktitle{International Conference on Internet-of-Things Design and Implementation (IoTDI '23), May 9--12, 2023, San Antonio, TX, USA}
\acmDOI{10.1145/3576842.3582325}
\acmISBN{979-8-4007-0037-8/23/05}

\begin{document}

\title[\sysname]{\sysname : Adaptive Privacy-Aware Reinforcement Learning for Sequential-Decision Making Human-in-the-Loop Systems}

\author{Mojtaba Taherisadr}
\email{taherisa@uci.edu}
\affiliation{%
  \institution{University of California Irvine}
  \city{Irvine}
  \state{CA}
  \country{USA}
  \postcode{92617}
}

\author{Stelios Andrew Stavroulakis}
\affiliation{%
  \institution{University of California Irvine}
  \city{Irvine}
   \state{CA}
  \country{USA}}
\email{sstavrou@uci.edu}

\author{Salma Elmalaki}
\affiliation{%
  \institution{University of California Irvine}
  \city{Irvine}
   \state{CA}
  \country{USA}}
\email{salma.elmalaki@uci.edu}

\renewcommand{\shortauthors}{Taherisadr \emph{et. al.}}

\input{00_abstract}


\vspace{-3mm}
\vspace{-2mm}


\maketitle

\input{01_introduction}

\input{02_relatedwork}
\input{03_rl}

\input{04_threatmodel}
\input{05_smarthome}

\input{06_smartlearning}
\input{08_conclusion}

\vspace{-3mm}
\section{Acknowledgment}
\vspace{-1mm}
This research was partially supported by NSF award \# CNS-2105084.


\appendix



\end{document}

%% file: 00_abstract.tex
\vspace{-2mm}
\begin{abstract}
Reinforcement learning (RL) presents numerous benefits compared to rule-based approaches in various applications. Privacy concerns have grown with the widespread use of RL trained with privacy-sensitive data in IoT devices, especially for human-in-the-loop systems. On the one hand, RL methods enhance the user experience by trying to adapt to the highly dynamic nature of humans. On the other hand, trained policies can leak the user's private information. Recent attention has been drawn to designing privacy-aware RL algorithms while maintaining an acceptable system utility. A central challenge in designing privacy-aware RL, especially for human-in-the-loop systems, is that humans have intrinsic variability, and their preferences and behavior evolve. The effect of one privacy leak mitigation can differ for the same human or across different humans over time. Hence, we can not design one fixed model for privacy-aware RL that fits all.  
To that end, we propose \sysname, an adaptive approach for privacy-aware RL, especially for human-in-the-loop IoT systems. \sysname provides a personalized privacy-utility trade-off depending on human behavior and preference. We validate the proposed \sysname on two IoT applications, namely (i) Human-in-the-Loop Smart Home and (ii) Human-in-the-Loop Virtual Reality (VR) Smart Classroom. Results obtained on these two applications validate the generality of \sysname and its ability to provide a personalized privacy-utility trade-off. On average, for the first application, \sysname improves the utility by $57\%$ over the baseline and by $43\%$ over randomization while reducing the privacy leak by $23\%$ on average. For the second application, \sysname decreases the privacy leak to $44\%$ before the utility drops by $15\%$.
\vspace{-2mm}

\end{abstract}

%% file: 01_introduction.tex
\vspace{-3mm}
\section{Introduction}

The emerging technologies of sensor networks and mobile computing give the promise of monitoring the humans' states and their interactions with the surroundings~\cite{mukhopadhyay2015wearable} and have made it possible to envision the emergence of human-centered design of Internet-of-Things (IoT) applications in various domains. This tight coupling between human behavior and computing enables a radical change in human life. 
By continuously developing a cognition about the environment and the human state and adapting the environment accordingly, a new paradigm for IoT systems provides the user with a personalized experience, commonly named Human-in-the-Loop (HITL) systems.  

The fundamental essence of designing HITL applications is learning the best adaptation to the environment, which is subjective to the human interaction and response to this adaptation which vary from one human to another~\cite{elmalaki2021fair}. Reinforcement Learning (RL) has proven to be adequate for monitoring human intentions and responses to provide such personalized adaptations~\cite{sadigh2017active, hadfield2016cooperative, elmalaki2022maconauto}. Multisample RL and adaptive scaling RL (ADAS-RL) 
can adapt to inter-and intra-human variability among humans and the changes in their response times under different autonomous actions~\cite{elmalaki2018sentio, ahadi2021adas}. 
Amazon has used personalized RL to adapt to students' preferences for adaptive class schedules~\cite{bassen2020reinforcement}. Advances in deep learning with RL have been used to decide which content to present to students at any given time based on their cognitive memory models~\cite{reddy2017accelerating}. 

This increasing adoption of RL-based models in various HITL applications has paved the way to reformulate the trained policies with constraints to address fairness~\cite{elmalaki2021fair}, risk-sensitivity~\cite{gagne2021two}, safety under exploration~\cite{hans2008safe}, and human variability~\cite{elmalaki2018sentio}. Adapting to the human often leads to systems where increased sophistication comes at the expense of more privacy weaknesses. In particular, RL has the added benefit of adapting to human variations to provide a personalized experience. However, privacy concerns are raised since the optimal trained policy holds a tight correlation between the human private state and the adaptation actions provided by the RL-based HITL system. For example, a smart NEST thermostat can automatically turn on and off the HVAC equipment based on users' presence or domestic activity~\cite{balaji2013sentinel}. 
Such coupling between human behaviors and decisions taken by the HITL system can open a side channel, leaking sensitive information about users' daily behavioral patterns. In particular, a malicious eavesdropper can infer a user's private information only by monitoring time-series data of the adaptation actions~\cite{erdemir2020privacy, elmalaki2022vindico}.

One of the critical challenges in designing HITL systems stems from the fact that the system's utility might be at odds with human expectations and privacy-preserving needs. While previous work in the literature addressed the problem of privacy leaks in learning-based adaptation engines, especially within RL-based models, through studying the fundamental privacy-utility trade-off~\cite{zhang2022privacy}, 
HITL systems hold a different challenge. In particular, we argue that privacy leak mitigation techniques should not be oblivious to the fact that there are intrinsic human variations. One privacy-aware RL model may cause severe degradation in the system utility from one human's perspective, while the same model can have acceptable utility for another human. This stems from the fact that every human's response, behavior, and interaction with the HITL application is different (\textbf{inter-human variation}). Moreover, human behavior may change over time, so even if we design an RL-based model that preserves the privacy of a particular human, it may not be adequate after some time as the human changes (\textbf{intra-human variations}). Hence, the concept of one privacy-aware model that fits all is inadequate for HITL systems. 

In this paper, we propose an \textbf{adaptive privacy-aware algorithm for RL-based models to provide personalized privacy-preserving human-in-the-loop IoT systems}. We borrow from the established theoretical underpinning of RL and information theory to formulate the problem as a sequential decision-making problem that maximizes the system utility with a personalized tunable regularizer that limits private information leakage due to human adaptation. 
We evaluate our proposed algorithm on two HITL IoT applications; the first one in the domain of smart home and the second one in the domain of smart classroom using Virtual Reality (VR). 

%% file: 02_relatedwork.tex
\vspace{-3mm}
\section{Related work and contribution}\label{sec:relatedwork}

Privacy has been a matter of concern for decades~\cite{petrescu2018analyzing}. Indeed, a plethora of work in the literature addresses privacy leaks and mitigation using a multitude of approaches. Game-theoretical approaches have been used to formulate an objective function that maximizes the utility and minimizes the privacy leaks~\cite{jin2017tradeoff}. Data encryption has been proposed to mitigate side-channel attacks on the communication links between the edge and cloud services~\cite{mishra2022secure}. 

\textbf{Our motivation behind the focus on RL methods stems from the following two fundamental properties:}
\begin{itemize}[leftmargin=*, topsep=0pt, noitemsep]
    \item \textbf{Computational complexity and scalability:} RL enjoys favorable computational scalability compared to other techniques especially game theoretic approaches. In particular, several game-theoretic approaches for sequential decision-making are known to be intractable~\cite{fraenkel2004complexity}.
    \item \textbf{Generalizability:} RL enjoys a unique ability to directly model the impact of taken decisions, leverage temporal feedback in learning, and improve the decision-making policy performance for a wide set of systems which is particularly important for HITL systems. 
\end{itemize}

\vspace{-3mm}
\subsection{Privacy-preserving RL} 
Various aspects of privacy-preserving RL problems have been considered and tackled, such as online learning with bandit feedback~\cite{malekzadeh2020privacy}, linear contextual bandits~\cite{garcelon2022privacy},
and deep RL (DRL)~\cite{pan2019you}. 
Garcelon et al.~\cite{garcelon2021local} formulated an algorithm that guarantees regret and privacy for the tabular setting. In the continuous state context, Wang et al.~\cite{wang2019privacy} developed a variant of Q-learning that can find a policy where the reward function satisfies the differential privacy constraints. 

The work by Erdemir et al.~\cite{erdemir2020privacy} studied the privacy-utility trade-off (PUT) in time-series data sharing. Existing approaches to PUT mainly focus on a single data point; however, temporal correlations in time-series data introduce new challenges. Methods that preserve privacy for the current time may leak a significant amount of information at the trace level as the adversary can exploit temporal correlations in a trace. They considered sharing the distorted version of a user's true data sequence with an untrusted third party. 

Liu et al.~\cite{liu2021deceptive} worked on the privacy of the reward function in RL systems. They tried to make it difficult for an observer to determine the reward function used. They presented two models for privacy-preserving reward. These models are based on dissimulation – a deception that `hides the truth.' They evaluated their models both computationally and via human behavioral experiments. The assumption in this study and other RL-based studies considering HITL systems is that the human state and available actions are finite and limited; otherwise, the implementation would be infeasible.

In this paper, we build upon work in the literature that exploited the correlation in the time-series data between the human state and adaptation action in RL with the assumption that HITL has finite and limited human states and adaptation actions. However, we differ from the work in the literature by arguing that the RL preserving algorithms should be adaptive to human variability in HITL IoT systems.

The closest to our approach is the work proposed by Cundy et al.~\cite{cundy2020privacy}. They proposed a regularizer based on the mutual information (MI) between the sensitive state and the actions at a given time step for sequential decision-making. They use an upper bound as an estimation of MI to guarantee that the policy derived from an RL algorithm satisfies the privacy constraint. They mathematically prove the correctness of the algorithm and then implement it on publicly available real-world data sets. 
The main difference between our proposed work (\textbf{\sysname}) and their approach is the adaptability of the privacy policy to inter-human and intra-human variations. In our experiments, we illustrate that using a constant upper bound privacy constraint, without considering the human variations, cannot efficiently satisfy a personalized privacy-utility trade-off in the HITL IoT systems.

\vspace{-4mm}
\subsection{Paper Contributions}
In this paper, we focus on sequential decision-making human-in-the-loop IoT systems with finite state/action pairs. In particular, we aim to address the potential privacy concerns that arise from sequential decision-making systems that interact with humans whose behavior and preference vary across time. 
Our contributions can be summarized as follows: 
\begin{itemize}[noitemsep,leftmargin=*, topsep=0pt, ]
    \item Designing \textbf{\sysname} - a privacy-aware RL-based algorithm for sequential decision-making HITL IoT systems that mitigates the privacy leaks adaptively based on human variability. 
    \item Providing general design parameters in our proposed \sysname algorithm that can be tuned based on the application domain.
    \item Implementing the proposed \sysname algorithm on two different HITL IoT systems in the domains of a smart home (simulation) and smart classroom with Virtual Reality (VR) (real-world). 
    \item Personalizing the trade-off between privacy mitigation and the application's utility.
\end{itemize}

%% file: 03_rl.tex
\vspace{-2mm}
\section{Human Modeling in Reinforcement Learning }\label{sec:humanmodelRL}
In the standard RL framework, a learning agent continuously interacts with an environment. The agent selects an action based on the current environment state, and the environment responds to this action by presenting a new state to the agent. This response is in the form of a feedback reward presented to the agent. The agent seeks to maximize the reward over time through its sequential decisions of actions~\cite{sutton2018reinforcement}. 
More formally, an RL agent interacts with an environment modeled as the Markov Decision Process (MDP) over a series of time steps $t \in \{0, 1, 2, ...\}$. At each time step, the RL agent takes action $a_t \in \mathcal{A}$ based on the current environment state $s_t \in \mathcal{S}$ and receives a reward $r_{t}: \mathcal{S}\times\mathcal{A} \rightarrow \mathbb{R} $ in the same time step\footnote{Some RL convention expresses the reward for action $a_t$ at time step $t$ in the next time step $r_{t+1}$.}. The dynamics underlying the environment can be described as an MDP with state-to-state transition probabilities, 
$p(s'|s,a) \doteq  Pr\{\mathcal{S}_{t+1} = s'|\mathcal{S}_t = s, \mathcal{A}_t = a\}$ 
and expected rewards for state-action pairs as: 
$r(s,a) \doteq  \mathbb{E} \{\mathcal{R}_{t} |\mathcal{S}_t = s, \mathcal{A}_t = a\}.$
Through repeated interaction with the environment, the agent tries to learn a state-action policy, $\pi(s, a) \doteq Pr\{\mathcal{A}_t = a|\mathcal{S}_t = s\}$
that maximizes the estimated reward over time. In the special case of deterministic policy, $\pi(s) \doteq a$ with probability$=1$ for $\mathcal{S}_t=s$.

\vspace{-2mm}
\subsection{Human as a Markov Decision Process}\label{sec:mdp}
Unique to the HITL systems is the integration of humans with the environment. Modeling the human in a way that captures the change in behavior and preference is an open, challenging research question. Borrowing up from the psychology literature, the behavior of the changes in the human decision historically was modeled through the expected utility theorem (EUT)~\cite{morgenstern1953theory}, which is based on an axiomatic framework defined as \emph{completeness}, \emph{transitivity}, \emph{independence}, and \emph{continuity}. Human is said to be rational if these four axioms hold. However, the EUT-based models have shown that these axioms are \emph{unrealistic} and that human decisions tend to deviate from the axioms of the EUT~\cite{tversky1979analysis}. Another approach to modeling the human is using the Partially Observed Markov Decision Process (POMDP)
based on the fact that even with advanced sensing technology, the actual human state can not be measured~\cite{rosenthal2011modeling}. Although POMDP aims to capture the human state's uncertainty, POMDP-based RL algorithms are computationally intractable, hindering their practical use~\cite{murphy2000survey}. 
Hence, in this work, we model the change in the human state as a Markov Decision Process (MDP) with unknown transition probabilities $p(s'|s, a)$. This uncertainty in the transition from one state to another can model the uncertainty and variability in the human state, which is essential in HITL systems.

\vspace{-3mm}
\subsection{Q-learning Reinforcement Learning}
Learning the optimal policy $\pi(s, a)$ ---action per state that maximizes the total reward---when the transition probabilities of the MDP model are unknown can be solved using RL. By applying an action in a particular state and observing the next state, the RL converges to the optimal policy that maximizes the reward function. This type of RL technique is called the Q-learning algorithm. The Q-learning algorithm assigns a value for every state-action pair. For each state $s$, the Q-learning algorithm chooses an action $a$ (among the set of allowable actions) according to a particular policy.
After an action $a$ is chosen and applied to the environment, the Q-learning algorithm observes the next state $s'$ of the environment and updates the \texttt{q-value} of the pair $(s, a)$ based on the observed reward $r(s, a)$ as follows: 
\begin{equation}\label{eq:qupdate}
\setlength\abovedisplayskip{0pt}
\belowdisplayshortskip=-1pt
Q(s, a) \leftarrow Q(s, a) +  \alpha[r(s,a) + \gamma \max_a Q(s', a) - Q(s, a)] 
\end{equation}

The hyperparameters $\gamma$ and $\alpha$ are known as the discount factor and the learning step size, respectively. 
To choose an action, $a$ at each state $s$, an $\epsilon$-greedy policy can be adopted. In the $\epsilon$-greedy policy, the RL agent chooses the action that it believes has the best long-term effect with probability $1-\epsilon$, and it picks an action uniformly at random; otherwise. In other words, at each time step, the RL agent flips a biased coin and chooses the action with the maximum \texttt{q-value} with probability $1-\epsilon$ or a random action with probability $\epsilon$. This hyperparameter $\epsilon$ (also known as the exploration vs. exploitation parameter) controls how much the RL agent is willing to explore new actions that were not taken before versus relying on the best action that has been learned so far. By updating $Q(s, a)$, it is guaranteed that the optimal policy $\pi$ will converge to a deterministic action $a$ per state $s$ that provides the maximum reward $r(s, a)$ in a finite time steps $T$~\cite{sutton2018reinforcement}.

%% file: 04_threatmodel.tex
\vspace{-2.5mm}
\section{Threat model in RL-based HITL IoT}\label{sec:threatmodel}
\begin{figure}[!t]
 \centering\setlength{\belowcaptionskip}{-15pt}
 \includegraphics[width=0.5\columnwidth]{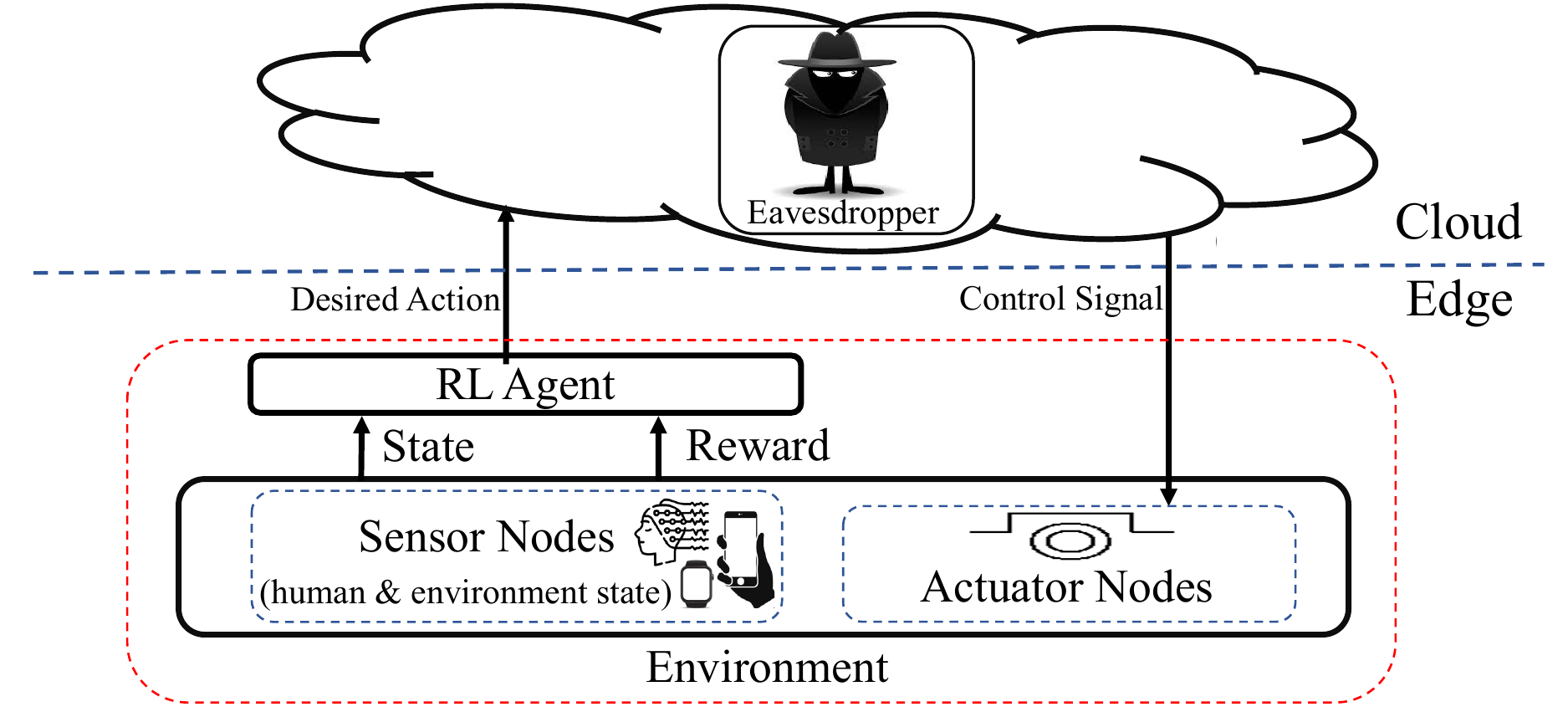}
  \caption{Threat Model for RL-based HITL IoT application.  } 
  \label{fig:attack_vector_big}
\end{figure}

Recent advances in edge devices’ power and memory capabilities paved the way to accomplish  relatively intensive computing on the edge. Accordingly, in \sysname, we can assume that the edge layer can handle relatively intensive data processing, including raw data processing and running an RL algorithm. 
Furthermore, synchronizing many IoT applications, especially in ubiquitous environments, requires a central decision-making server at a cloud-based level. Hence, the edge does not send the control actions directly to the environment and has to share it first with the cloud for other constraints that the cloud may need to enforce, such as synchronizing multiple IoT applications. This IoT edge-cloud computing model is a typical architecture for many pervasive and ubiquitous IoT applications~\cite{bovornkeeratiroj2020repel,papst2022share,stirapongsasuti2019decision}. 
A pictorial figure for our proposed threat model is shown in Figure~\ref{fig:attack_vector_big}, which can be summarized as follows: 
\begin{itemize}[topsep=0pt, noitemsep]
\item An HITL IoT application collects information from the environment and the human interacting with it through multiple sensors on edge devices. 
\item An RL agent --- as explained in Section~\ref{sec:humanmodelRL} --- runs on the edge to infer the human and environment state and recommends the desired adaptation action based on human preference and behavior. 
\item Only the desired action recommended by the RL agent is propagated to a cloud-based server. In particular, the raw data from the sensors at the edge and the inferred human or environment states are not shared with the cloud. The edge is a trusted entity.
\item The cloud elects the appropriate control signals that can be based on other enforced constraints and sends them to the actuator nodes (edge devices) to adapt to the environment.
\end{itemize}

\noindent Hence, based on this threat model, our attack vector is as follows:

\begin{itemize}[topsep=0pt, noitemsep]
    \item An eavesdropper is mounted in the cloud and has no access to the edge devices (sensors or actuators) or the communication channel between the edge and the cloud.
    \item The eavesdropper has access to the time-series data of the desired actions in the cloud.
    \item The eavesdropper can run any machine learning model exploiting this time-series data.
    \item The eavesdropper has prior knowledge of the application domain. 
\end{itemize}

In this attack vector, even if the communication channel between the edge and the cloud may be secured (e.g., by encryption), the cloud still needs to decrypt the transferred packet --- which contains the desired action --- to elect the control signals that are sent back to the actuators at the edge for ubiquitous IoT applications. Hence, the eavesdropper can observe and record the time-series data of the desired actions of this IoT application.

Our goal in \sysname is to provide a guarantee that limits the ability of an eavesdropper to infer the private state, even with unlimited computational power and complete knowledge of the application domain. 

\vspace{-3mm}
\section{\sysname: Adaptive Privacy-aware RL}\label{sec:adaparl}\vspace{-1mm}
As explained in Section~\ref{sec:humanmodelRL}, the state in MDP holds information about the environment, including sensitive information about the human interacting with it, such as human mental state, location, gender, or behavioral activity. The fact that the optimal policy $\pi(s, a)$ is a function of the state-action pairs can open a side channel that leaks the private human state $s$. We formulate the problem of privacy-aware RL as a Markov decision process (MDP) that ensures privacy constraints. In particular, we aim to learn a policy $\pi(s, a)$ at the edge, which maximizes the cumulative reward while constraining the privacy leak of the sensitive private state to an eavesdropper mounted in the cloud. 

\vspace{-3mm}
\subsection{State-Action Dependence}\label{sec:stateactiondep}
We draw on the information theory literature and leverage \emph{mutual information} (MI) to quantify the amount of correlation (or dependence) between two random variables. In our problem, we use the MI between \emph{state} and \emph{action} to measure how certain an eavesdropper can infer a \emph{state} from observed \emph{actions}. MI provides a theoretical bound on the inference capability of \emph{any} learning algorithm. 
Generally speaking, the lower the MI between state and action is, the lower the accuracy of \emph{any} inference algorithm.
Push into one extreme; if the MI is zero, then \emph{no} algorithm can infer the state from monitored actions. 
Hence, we consider the MI measure $I_\pi(a_t; s_t)$ for a particular policy $\pi(s, a)$ as a quantifiable bound on the ability to obtain the highest amount of information on $s_t$ by observing $a_t$. In particular, the amount of information leaked to the eavesdropper is bounded by $I_\pi(a_t; s_t)$\footnote{Since we focus on HITL IoT applications, the number of states and actions are finite and limited as mentioned in Section~\ref{sec:relatedwork}. Hence, estimating the MI in this setup is tractable.}.

Hence, we formulate the privacy-aware policy RL problem as a regularized optimization problem that maximizes the cumulative reward while maintaining a bound on them $I_\pi(a_t; s_t)$. In particular, we add an adaptive regularizer that penalizes the reward through a constraint on the value of $I_\pi(a_t; s_t)$. Hence, the reward value $r(s,a)$ to calculate the \texttt{q-value} as explained in Equation~\ref{eq:qupdate} can be formulated as:

\begin{equation} 
\setlength\abovedisplayskip{0pt}
\belowdisplayshortskip=-1pt
r(s,a) = \begin{cases}
 \mathbb{E} \{\mathcal{R}_{t} |\mathcal{S}_t = s, \mathcal{A}_t = a\}, & \text{ if } I_{\pi}(a_t;s_t) < \lambda_t  \\

 (1-\zeta_t) \mathbb{E} \{\mathcal{R}_{t} |\mathcal{S}_t = s, \mathcal{A}_t = a\} - \zeta_t I_{\pi}(a_t;s_t), & \text{ otherwise}
\end{cases}
\end{equation}\\
Where $\zeta_t \in [0,1]$ is a design parameter regulating the trade-off between the privacy leak mitigation and the application utility, commonly known as the privacy-utility tradeoff (PUT). The constraint  $I_{\pi}(a_t;s_t) < \lambda_t$ is used to set a boundary on the rise of MI by penalizing the reward function. Unique to \sysname is that the boundary $\lambda_t$ is adaptive and is human and application dependent, as we will explain in Section~\ref{sec:app1} and ~\ref{sec:app2}.

\vspace{-2mm}
\subsection{Adaptive Privacy-Aware Constraint}\label{sec:adaprivacy}

It is worth mentioning that penalizing the reward with $I_{\pi}$ should be adaptive based on human variability. In particular, we make the following observations: 
\begin{itemize}[noitemsep, leftmargin=*, topsep=0pt]
    \item \textbf{High intra-human variability:} If the human behavior and preferences frequently change with no particular pattern, the reward received by the Q-learning agent will be a dynamic and time-varying reward. This means the optimal policy $\pi(s, a)$ will take more time to converge. Consequently, the value of $I_{\pi}$ will rise very slowly as the agent learns the personalized policy $\pi(s, a)$.
    \item \textbf{Low intra-human variability:} In contrast, the reward will be less dynamic if the human has repeated patterns with the same expected behavior and preference. Hence, the Q-learning agent will learn the personalized policy $\pi(s, a)$ in less time. Consequently, the value of  $I_{\pi}$ will rise quickly. 
    \item The maximum value of $I_{\pi}$ for a particular policy $\pi(s,a)$ depends on the time-series of the state $s_t$ and the action $a_t$ which are application dependent. 
\end{itemize}

Based on these observations, the upper bound $\lambda_t$ should be adaptive based on how the MI increases, which is correlated to human variability. In particular, in \sysname, we propose the following adaption strategy to find the appropriate $\lambda_t$.

As the agent learns the optimal policy $\pi(s, a)$, it keeps track of a time series of $I_{\pi}(a_t;s_t)$. This time series is used to fit a higher-order polynomial function of degree $2$. Based on this fitted function, we can approximate the growth rate and the maximum value that the $I_{\pi}(a_t;s_t)$ can reach. The upper bound $\lambda_t$ is then set at a particular percentage of this maximum value. This percentage -- which we call $\lambda_{percent}$ -- is a design parameter, as we will show in our evaluation. Hence, the constraint on $I_{\pi}(a_t;s_t)$ is not based on an absolute value but rather on how the agent learns, which is human-dependent (inter-human variability). Intuitively, this means that we allow the agent to learn to provide acceptable utility before it is penalized through the privacy constraints $I_{\pi}(a_t;s_t) < \lambda_t$. Indeed, the fitted curve is corrected over time through more interaction with the environment and to track any changes in human behavior. Hence, the value of $\lambda_t$ is also time-varying depending on the changes in human behavior (intra-human variability)

\vspace{-3mm}
\subsection{\sysname Algorithm}
Algorithm \ref{algorithm1} summarizes the general \sysname algorithm. Indeed the reward function $\mathcal{R}(s_t, a_t)$ in \sysname algorithm is application dependent. Moreover, it is worth mentioning that the MI value ($I_t$) in the algorithm is calculated over a time series of states ($\mathcal{H}_{\mathcal{S}}$) and actions ($\mathcal{H}_{\mathcal{A}}$) which are based on the sequential decision-making of the RL-agent at every time step $t$. Hence, by using the reward shaping approach in the \sysname algorithm with the adaptive MI regularizer, the \sysname agent learns to consider the consequences of choosing an action $a_t$ at time step $t$ on the distribution of states $s$ at the future time steps, which is human dependent. Eventually, \sysname agent will choose future action $a_t$ that decreases the dependency on $s_t$ for a particular human. To evaluate the proposed \sysname algorithm, we design HITL RL-based IoT applications and show that the adaptation to human preferences is achieved. Afterward, we discuss the privacy leaks that may occur due to the RL-based adaptation, which is application dependent. 
Ultimately, we show how \sysname can mitigate privacy leaks and provide the privacy-utility trade-off (PUT) that varies across different humans. Accordingly, we discuss two IoT applications. The first one is in the domain of smart house (Section~\ref{sec:app1}), a simulation-based application to evaluate the different design parameters in a controlled simulated environment. The second one is in the domain of smart classrooms (Section~\ref{sec:app2}) using Virtual Reality (VR), which is a real-world experiment. 
Through providing these two different application scenarios, we aim to evaluate the applicability and adaptability of the  \sysname in different situations and on different people (inter- and intra-human variability) and to emphasize the generalizability of \sysname.

\setlength{\textfloatsep}{0pt}
\begin{algorithm}
\caption{\sysname algorithm}\label{alg:cap}
\begin{algorithmic}
\State \textbf{Q-Learning hyperparameters:} $\alpha$, $\gamma$, $\epsilon$
\State \textbf{adaPARL design parameters:} $\zeta$, $\lambda_{percent}$
\Require 
\Statex States  $\mathcal{S} = \{1, \ ... \ ,S_n\}$, Actions  $\mathcal{A} = \{1, \ ... \ ,a_n\}$
\Statex Reward function $\mathcal{R} = \mathcal{S} \times \mathcal{A} \rightarrow \mathbb{R} $
\Statex Transition function $\mathcal{T}: \mathcal{S} \times \mathcal{A} \rightarrow \mathcal{S}$
\Statex Privacy-Utility trade-off $\zeta \in [0, 1]$
\Statex Privacy mitigation upper bound $\lambda_{percent} \in [0, 1]$
\Statex Mutual Information Queue  $\mathcal{MI} = [] $ 
\Statex States History Queue  $\mathcal{H_{\mathcal{S}}} = [] $, Actions History Queue  $\mathcal{H_{\mathcal{A}}} = [] $ 
\Statex Learning rate $\alpha \in [0, 1]$,  $\alpha = 0.01$
\Statex Discounting factor $\gamma \in [0, 1]$, $\gamma = 0.001$
\Statex $\epsilon$-Greedy exploration strategy with decay  $\epsilon \in [0, 1]$, 
\Statex \quad\quad max $\epsilon = 0.9$, min $\epsilon = 0.1$, decay$ =0.01$

\Procedure {\sysname}{$\mathcal{S}$, $\mathcal{A}$, $\mathcal{R}$, $\mathcal{T}$, $\gamma$, $\alpha$, $\epsilon$, $\zeta$,  $\lambda_{percent}$} 
            \State Initialize $Q: \mathcal{S} \times \mathcal{A} \rightarrow \mathbb{R}$ with 0
            \State time sample $t=0$
            \State Observe initial state $s_t \in \mathcal{S}$
            \While{true}
                \State Apply $\pi(s)$ according to the exploration strategy:
                \State\qquad {\small{with probability $\epsilon$: $\pi(s) \gets$ choose $a \in \mathcal{A}$ at random,
                \State\qquad with probability $1-\epsilon$: $\pi(s) \gets \argmax_{a} Q(s, a)$)}}
                
               \State $a_t \gets \pi(s_t)$ \Comment{Choose a desired action}
                \State  push $a_t$ to $\mathcal{H_{\mathcal{A}}}$ and  push $s_t$ to $\mathcal{H_{\mathcal{S}}}$
                    \State calculate  $I_t(\mathcal{H_{\mathcal{S}}}; \mathcal{H_{\mathcal{A}}}) $ \Comment{Calculate Mutual Information}
                    \State  push $I_t$ to $\mathcal{MI}$ 
                    \State  $f(I_t) = $  polynomial function of degree 2 fitted to $\mathcal{MI}$ 
                    \State    $\lambda_t$ =  $f_{max}(I_t) \times \lambda_{percent}$
                    \If{$I_t$  $<$ $\lambda_t$}
                      \State $r(s_t,a_t) = \mathcal{R}(s_t, a_t)$ \Comment{Receive the performance reward}
                    \Else{} \Comment{Penalize based on the privacy leak}
                      \State $r(s_t,a_t) = (1-\zeta) \mathcal{R}(s_t, a_t) - \zeta I_t(a_t;s_t)$
                    \EndIf
                    \State $s'_t \gets \mathcal{T}(s_t, a_t)$ \Comment{Observe the next state}
                    \State $Q(s_t, a_t) \leftarrow Q(s_t, a_t) +  \alpha[r(s_t,a_t) + \gamma \max_a Q(s'_t, a) - Q(s_t, a_t)] $
                    \State $s_t \gets s_t'$
            \EndWhile
        \EndProcedure
\end{algorithmic}
\label{algorithm1}
\end{algorithm}

\vspace{-2mm}
\subsubsection{Generalizability of \sysname}
 The generalizability of \sysname comes from the two designed parameters, $\zeta_t$ and $\lambda_t$. As explained in Section \ref{sec:adaprivacy}, the $\lambda_t$ is set at a particular percentage of the maximum value of $I_{\pi}(a_t;s_t)$. This percentage -- which we call $\lambda_{percent}$ -- is a design parameter correlated to human behavioral variations and provides the notion of personalized adaptation of privacy leak mitigation. Moreover, the privacy-utility trade-off can be tuned through the parameter $\zeta_t$, as explained in Section~\ref{sec:stateactiondep}. Accordingly, the \sysname algorithm can be used and implemented in various applications that need tuning of the privacy-utility trade-off. 

%% file: 05_smarthome.tex
\vspace{-2mm}
\section{Application 1: Human-in-the-Loop Smart Home- A Thermal System}\label{sec:app1}
Recent work in the literature targets human-in-the-loop smart heating, ventilation, and air conditioning system (HVAC) while trying to assist human satisfaction~\cite{jung2017towards}. RL has been proposed to adapt the HVAC set-point based on human activity~\cite{elmalaki2021fair}. A human-in-the-loop HVAC system should take the human state and preferences into the computation loop while calculating the HVAC set-point. For example, the human body temperature decreases when the human goes to sleep,
while the body temperature increases when the human exercises 
and with stress and anxiety.
Monitoring the human state,
sleep cycle, and 
physical activity 
are all possible with IoT edge devices~\cite{likamwa2013moodscope}. While the main purpose of this section is to evaluate the privacy-utility trade-off provided by \sysname, we first describe the environment design and the RL agent used for this application.

\vspace{-3mm}
\subsection{System Design \& Implementation}\label{sec:app1design}
\emph{\textbf{Environment Design: }} We simulated 
thermodynamic model of a house that considers the house's geometry, the number of windows, the roof pitch angle, and the type of insulation used. The house is heated by a heater with an airflow of temperature $50^{\circ}c$ and cooled by a cooler with an airflow of temperature $10^{\circ}$. A thermostat allows fluctuation of $2.5^{\circ}c$ above and below the desired set-point, specifying the temperature that must be maintained indoors~\cite{MATLABther}. The desired set-point is controlled by an external controller that runs the proposed \sysname algorithm.

\emph{\textbf{Simulated Human Model: }} We model the humans as a heat source with heat flow that depends on the average exhale breath temperature ($EBT$) and the respiratory minute volume ($RMV$)~\cite{elmalaki2021fair}. 
The $RMV$ is the product of the breathing frequency ($f$) and the volume of gas exchanged during the breathing cycle, which is highly dependent on human activity. For example, $RMV \approx 6$ $l$/$m$ when the human is resting while $RMV \approx 12$ $l$/$m$ represents a human performing moderate exercise~\cite{carroll2006elsevier}. We simulated the behavior of three humans based on their activity. The human activity is simulated by different values of the $RMV$ ~\cite{carrollpulmonary2007} 
and the metabolic rate. 
We simulated four activity classes, including three in-home activities and a ``not at home'' state. Since some activities have close $RMV$ and their differences do not affect the primary goal of this study, we categorized the normal human activities inside the house into three groups. The three in-home activity categories arranged in ascending order of $RMV$ are sleeping, relaxed (sitting, standing, reading, and watching TV), and medium domestic work (washing dishes, cooking, and cleaning). We randomize the behavior by having different choices of activities in the same time slot to design different human daily behavior. We assume that the age/sex/time of day has no significance in the model. We extended the thermal house model by Mathworks~\cite{MATLABther} to include a cooling system and a human model\footnote{While there are more sophisticated simulators for smart houses and smart buildings that consider the energy consumption and the electric loads, such as EnergyPlus~\cite{gerber2014energyplus}, we opt for a simpler model of the thermal house to evaluate \sysname.}.

\vspace{-2mm}
\subsection{RL Design} \label{sec:app1RLdesign}\vspace{-1mm}
To adapt the HVAC set-point based on the human activity and thermal comfort level, we designed an RL  as described below.  
\vspace{-2mm}
\subsubsection{\textbf{State and Action Space}}
State Space $\mathcal{S}$: 
All combinations of four distinct human activities, $\mathcal{S}  = \{(act) : act \in [1,4]\}$, 
where $act$ is the current human activity as mentioned in Section~\ref{sec:app1design}.  
Action Space $\mathcal{A}$: A discrete value for the set-point within the temperature range [60,80] with a heater option or a cooling option:
$\mathcal{A} = \{a : a \in [60, 80]\}$.

\vspace{-2mm}
\subsubsection{\textbf{Designing the reward function}}
Reward $\mathcal{R}$: 
We use the Prediction Mean Vote (PMV) as an estimation for the
human thermal comfort~\cite{fanger1970thermal}. 
The scale of PMV ranges from $-3$ (very cold) to $3$ (very hot). According to ISO standard ASHRAE 55~\cite{handbook2009american}, a PMV in the range of $[-0.5, 0.5]$ for the interior space is recommended to achieve thermal comfort. Estimating the PMV score is calculated based on the knowledge of clothing insulation, the metabolic rate, the air vapor pressure, the air temperature, and the mean radiant temperature~\cite{fanger1970thermal}. We use a simple reward value based on the comfort value of the human dictated by the PMV. In particular, the comfortable thermal sensation $PMV = [-0.5, 0.5]$ receives positive and higher rewards, and the discomfort levels of PMV receive negative rewards. 
In practice, the PMV value can be estimated using edge devices, such as black globe thermometers~\cite{Globe}.

\vspace{-2mm}
\subsubsection{\textbf{Hyperparameters selection}}
We briefly list some of the hyperparameters in the design of the RL agent.
\begin{itemize}[noitemsep, topsep=0pt, leftmargin=*]
    \item \emph{\textbf{Discount factor $\gamma$:}} In Equation~\ref{eq:qupdate}, $\gamma$ determines how much the RL agent cares about rewards it receives in the distant future relative to the immediate reward. In our design, the \texttt{q-value} updates only when the indoor temperature reaches the selected set-point (the selected action $a$), independently of how long it takes for the indoor temperature to reach this set-point which is dictated by the thermal dynamics of the house. Hence, a low discount factor $\gamma=0.001$ is selected.
    \item \emph{\textbf{Exploration vs. Exploitation $\epsilon$:}} Exploration is critical due to the inter- and intra-human variability. Once the agent has the appropriate information through interaction with the HITL environment, it is better to lower the exploration rate. Hence, every time \texttt{q\_value} is updated, as explained in Equation~\ref{eq:qupdate}, we gradually lower $\epsilon$ following an exponential decay of $0.01$.
\end{itemize}

\vspace{-3mm}
\subsection{Human-in-the-Loop RL Adaptation}\label{humaninthelp}
We simulated three different humans. Each human has been designed to have different life patterns. $H_1$'s life pattern follows an organized pattern and is repeated weekly with limited randomness. $H_3$ has a more random life pattern meaning that activities do not follow a specific daily or weekly routine, and it contains numerous unexpected changes. $H_2$ contains medium randomness, which in terms of randomness stands in between $H_1$ and $H_3$. For example, a pictorial image for the behavioral pattern of $H_2$ is shown in Figure~\ref{fig:cluster1}, where the human behavior alternates between 4 main activities (sleeping, not at home, domestic activity, and relaxed). 
We run the simulator for $8000$ time step ($T_s = 6 min$ simulation time).   
To elaborate on how the designed RL-agent learned the best set-point per human activity, we split the RL-agent actions into three different plots (for clarity) based on the human activity at a particular time step, as shown in Figure~\ref{fig:setpointconvergence} for $H_2$.  
In particular, as Figure~\ref{fig:setpointconvergence} presents, approximately after $350$ hours ($15$ simulated days), the random selection of the set-point decreased, indicating that the RL-agent started to learn the appropriate set-point for this activity.

\subsection{Information Leak}\label{sec:privacyleak}
\vspace{-2mm}
We evaluate the threat of the private information leak in this application by assuming an eavesdropper who can monitor the time series of actions decided by the RL agent. This is possible assuming a smart thermostat system that uses a cloud-based service, such as NEST~\cite{nest} with a mounted spyware eavesdropper in the cloud. 
State in the RL model (human activity) can be determined from sensor nodes inside the house or wearable devices. RL model runs at the edge (such as a mobile phone). The RL model sends the desired set-point (action) to the cloud engine. In this model, the attacker is located in the cloud engine, where the control signal is generated and sent to the HVAC thermostat to control the environment (house).

\begin{figure}
{\begin{minipage}[b]{0.5\textwidth}
\centering
\includegraphics[scale=0.42]{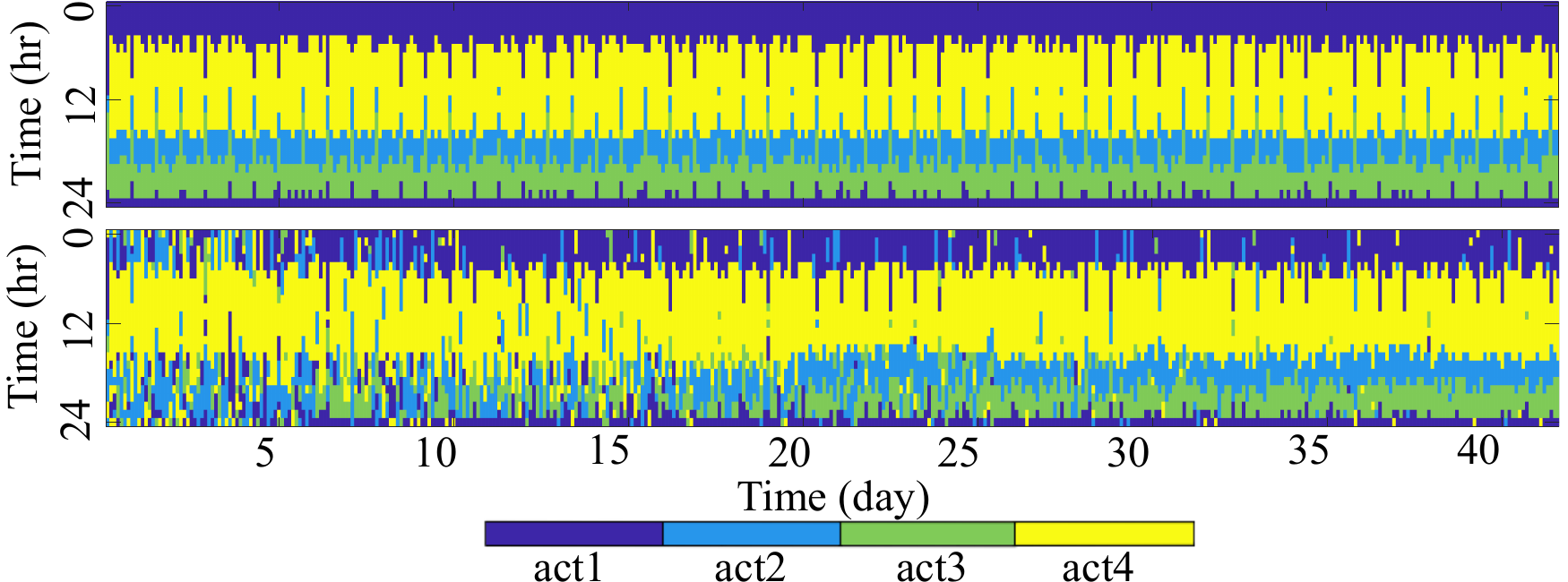}\\
\caption{Top: A human daily activity ($H_2$) inside a house across the $24$ hours of the day for $40$ days. act1, act2, act3, and act4 are sleeping, not at home, domestic activity, and relaxed, respectively. Bottom: The clustering of the RL actions (set-points) as seen by the eavesdropper with no access to the actual human behavior ground truth.}
\label{fig:cluster1}
\end{minipage}}\hspace{2mm}
\raisebox{5 ex}{\begin{minipage}[b]{0.45\textwidth}
\centering
\includegraphics[scale=0.80]{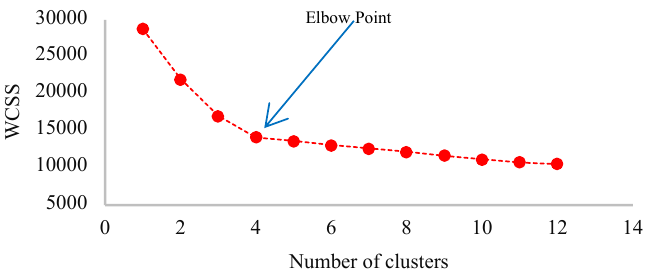}\\
\caption{WCSS (Within-Cluster Sum of Square) vs. the number of clusters. The elbow point presents the most efficient number of clusters.}
\label{fig:elbowreward}
\end{minipage}}%
\end{figure}

Since the eavesdropper has no prior knowledge of human behavior inside the house, unsupervised learning techniques can be used to infer the hidden patterns. For example, if the eavesdropper uses a clustering algorithm, such as $K$-means, we can show that sensitive information, such as occupancy and sleeping time, can be leaked. Since the eavesdropper has no prior knowledge of human activity, the number of clusters is unknown. Hence, a common technique an eavesdropper can do is to use the elbow point to determine the best number of clusters. Figure~\ref{fig:elbowreward} shows the elbow point result. Four clusters are the most dominant result for the clustering numbers, which equals the actual number of human activities in the simulation model.

Accordingly, we compare the ground truth (the actual behavior of $H_2$ human in the simulation model) and the clustering results by an eavesdropper (4 clusters) for each day respectively in Figure~\ref{fig:cluster1} top and bottom, respectively. As Figure~\ref{fig:cluster1} (bottom) illustrates, the eavesdropper can cluster the actions (set-point) meaningfully, which is correlated with the pattern of human activity. In this case, the eavesdropper can achieve a clustering accuracy of $86\%$. This clustering results by the eavesdropper can show that approximately after $15$ days (as  illustrated in Figure~\ref{fig:setpointconvergence}), there are some information leaks on the behavioral pattern of the human. Indeed the exact human activity is not inferred. However, we show here that the eavesdropper can infer some behavioral pattern of the human with the knowledge of the application domain, such as when the human most likely goes to sleep or when the human leaves the house or any changes in the normal human daily behavior.  

\vspace{-3mm}
\subsection{Privacy Leak Mitigation with \sysname }\label{daPARL}
We evaluated \sysname in adaptively mitigating the privacy leak based on human behavior while providing a privacy-utility trade-off. In particular, we compare \sysname with two approaches. The first is a naive approach of randomizing some actions propagated to the cloud by adding noise (sending random action instead of the RL decided action). The second approach is what we call a ``fixed privacy'' mitigation approach~\cite{cundy2020privacy} which will be the baseline to which we compare \sysname. In all three approaches; randomization, fixed privacy mitigation, and \sysname; the main objective is to reduce the MI to limit the eavesdropper's ability to infer the correlation between the state and the action pairs independently of the machine learning algorithm used by the eavesdropper. Figure~\ref{fig:allbigexp1} shows the summary for this comparison across different human behavior (as explained in Section~\ref{humaninthelp}) which we explain in detail in this section.

\subsubsection{\textbf{Mitigation 1: Randomization}}\label{sec:randomization}
\begin{figure}[!t]
\raisebox{2.5 ex}{\begin{minipage}[b]{0.485\textwidth}
\centering
\includegraphics[scale=0.5]{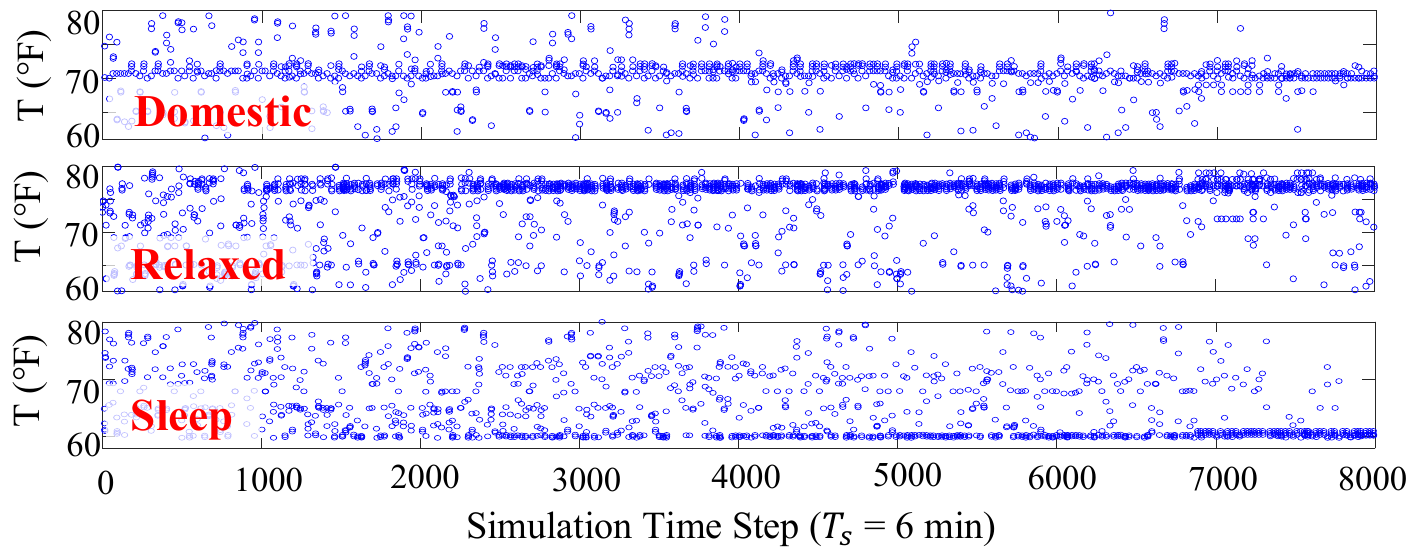}
\caption{RL Actions (set-point) divided into three plots. Set-point converges to $\approx 60^\circ F$ for sleeping, $\approx 77^\circ F$ for relaxed, and $\approx 71^\circ F$ for domestic activity.}
\label{fig:setpointconvergence}
\end{minipage}}\hspace{2mm}
\begin{minipage}[b]{0.48\textwidth}
\centering
\includegraphics[scale=0.41]{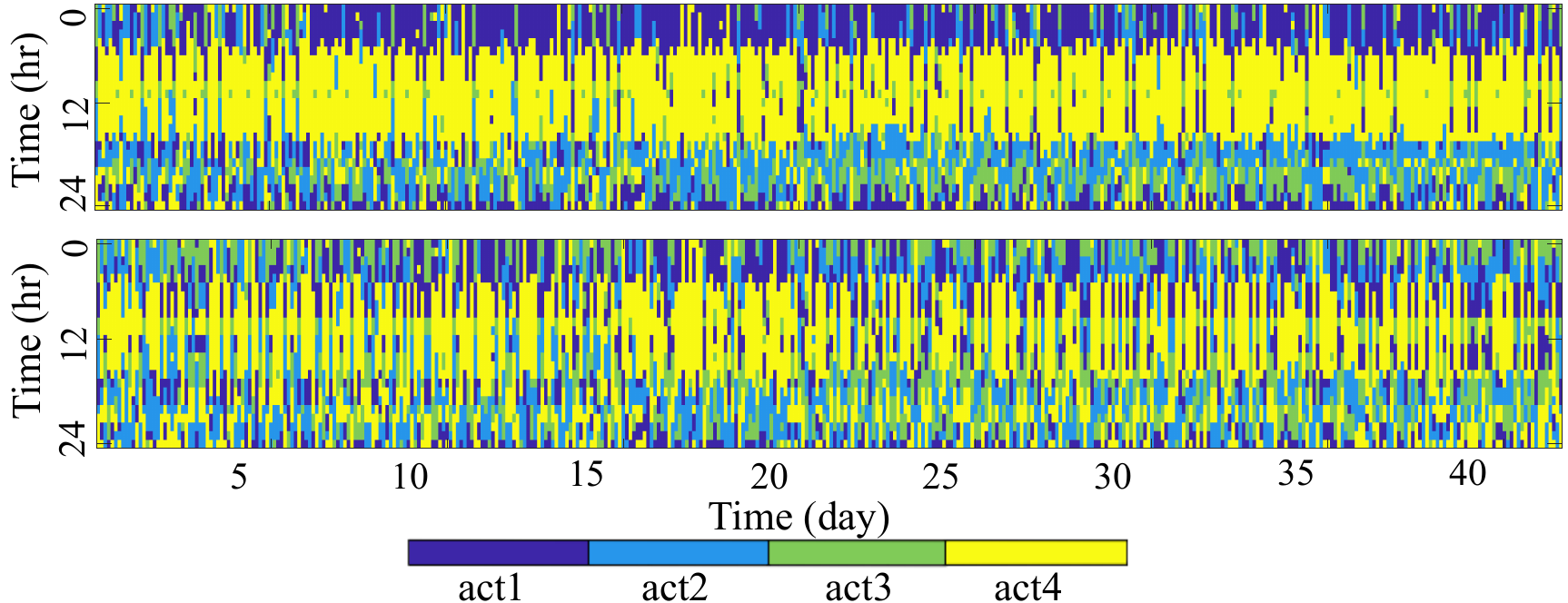}\\
\caption{Clustering before and after using mitigations. Top: clustering results with randomization mitigation with $p=0.5$. Bottom: clustering result of \sysname with $\lambda_{percent}=80\%$ as seen by the eavesdropper with no access to the actual human behavior.}
\label{fig:cluster2}
\end{minipage}
\end{figure}

A biased coin with a probability $p$ is used to decide whether to randomize the currently selected action $a_t$. In particular, if $p = 0.5$, then $50\%$ of the selected actions are masked through randomization before being sent to the cloud. In Figure~\ref{fig:cluster2} (Top), we show the clustering results as seen by an eavesdropper in the cloud with $p = 0.5$. The accuracy of clustering dropped to $65\%$ compared to $86\%$ before randomization. 
Figure~\ref{fig:pmv_bfr_aftr_rand} (left) presents the effect of adding randomization on the human comfort level $PMV$. After randomization, the $PMV$ histogram expands toward values outside the acceptable range of human thermal comfort. As expected, adding randomization decreases the eavesdropper's ability to predict the human's daily behavior. However, it comes at the cost of reducing the application utility measured by the $PMV$, where human experiences more uncomfortable thermal comfort moments.

We evaluated different values of $p$ and their effect on the MI on different human behavior. 
As seen in Figure~\ref{fig:allbigexp1} (row 2), the MI before and after adding randomization (with $p=0.5$) to the RL actions across $8000$ simulation time steps, where each time step is equivalent to a simulated $6$-minute in the system model we described in Section~\ref{sec:app1design}. As expected, adding randomization limits the MI. In particular, the MI reaches $1.45~bits$ on average for the simulated $3$ humans without mitigation. By adding randomization, the MI is decreased to a value less than $1~bits$ on average\footnote{The unit of MI value depends on the base of the logarithm. If base $2$ is used, MI is measured in bits.}.

To evaluate the privacy-utility trade-off using this approach, we use the standard deviation (STD) of the $PMV$ as a measurement of utility. In particular, as the value of the STD of the $PMV$ increases, it indicates a low utility (more spread of $PMV$ value). We plot the clustering accuracy as an indication of privacy leak vs. the STD of the $PMV$ for different randomization values $p$ as shown in Figure~\ref{fig:allbigexp1} (row 3). 
Increasing the randomization leads to better mitigation of the privacy leak. However, this privacy leak mitigation costs higher STD for $PMV$, meaning that human experiences discomfort, indicating low application utility. While the clustering accuracy drops to approximately $65\%$ at $p=0.5$, the STD of the $PMV$ exceeds $1$, which means $PMV$ values are more than $1$ or less than $-1$, indicating high discomfort levels of the $PMV$. We chose not to increase $p$ more than $0.5$ due to the increase in the STD of $PMV$ beyond $1$.

\vspace{-2mm}
\subsubsection{\textbf{Mitigation 2: }\sysname}\label{adaparlintra}
\begin{figure}
 \centering \includegraphics[width=0.9\textwidth,scale=0.425]{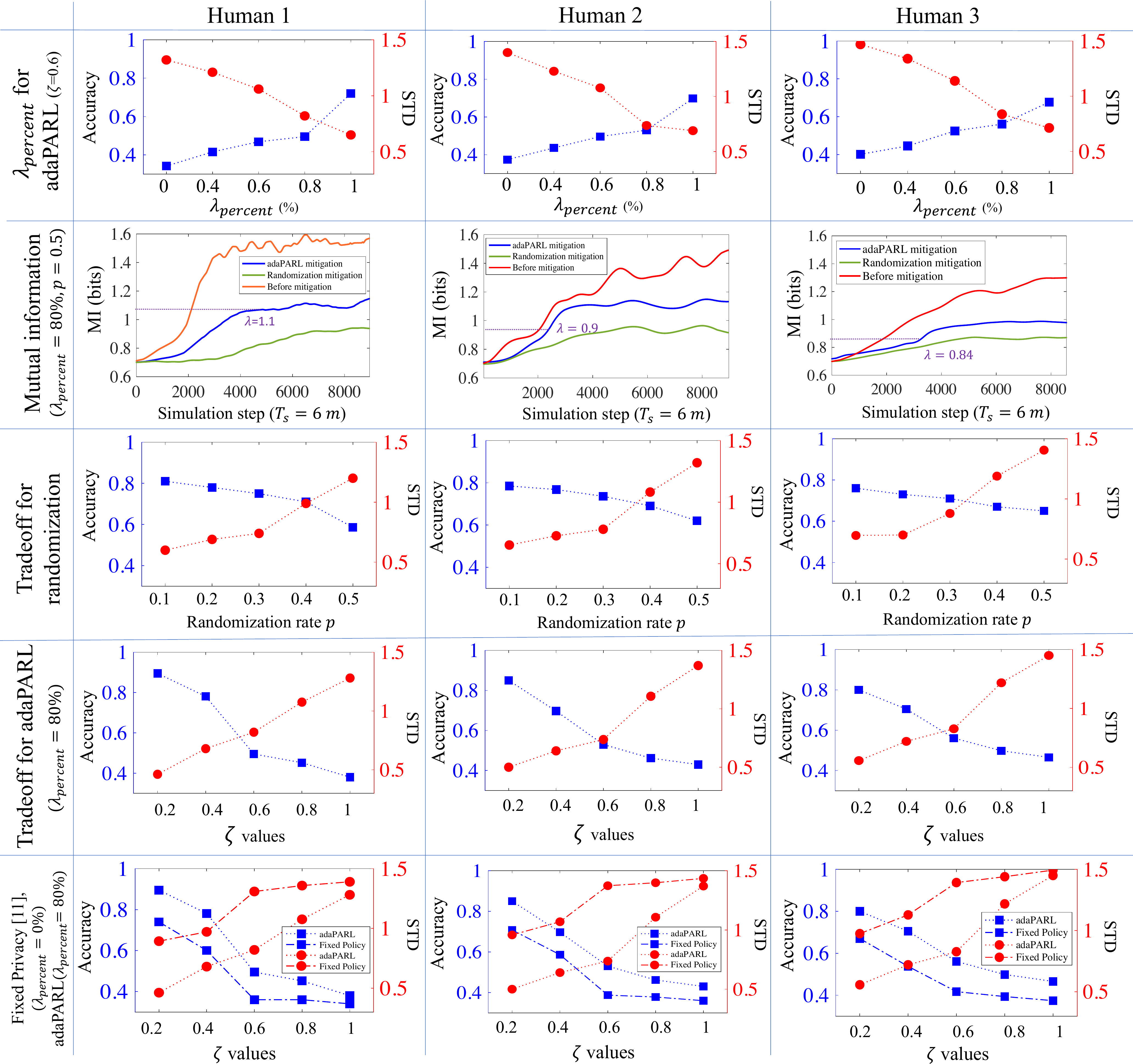}
  \caption{
  Proposed \sysname algorithm parameters analysis and privacy vs. utility trade-off. The first row provides privacy vs. utility trade-off by considering different values of $\lambda_{percent}$ for standard deviation (STD) of $PMV$ on the right vertical axis and clustering accuracy on the left vertical axis. The second row depicts the MI growth rate for different mitigation algorithms. The third row presents the privacy vs. utility trade-off for the randomization mitigation approach. The fourth row illustrates the privacy vs. utility trade-off for \sysname algorithm. On the right vertical axis, it presents the STD of the $PMV$ and clustering accuracy on the left vertical axis across different values of design parameter $\zeta$ on the horizontal axis. The last row shows the privacy vs. utility trade-off comparison between \sysname and baseline algorithm proposed in~\cite{cundy2020privacy}. 
  }\label{fig:allbigexp1}
\end{figure}

While the first approach of adding randomization can achieve acceptable privacy leak mitigation, we argue that this mitigation has to be adaptive and not oblivious to human behavior. 
We compare the MI using \sysname with the randomization approach in the same Figure~\ref{fig:allbigexp1}~(row 2) using $\lambda_{percent} = 80\%$. The MI is reduced and limited to $\sim 1~bits$ on average for the three humans. As expected, the value of $\lambda$ differs for the three humans. Across the three humans, the MI before mitigation is different due to the different behavioral patterns (\textbf{inter-human variability}). Behavior patterns represent the complexity and non-uniformity of human activity in terms of repetition and order during the day. A more complex life pattern results in MI that grows slower ($H_3$), and the RL agent requires a longer time to learn the policy $\pi(s, a)$ in contrast with a more repetitive behavior ($H_1$). Hence, with $\lambda_{percent}=80\%$, the value of $\lambda$ differs depending on human behavior to provide adequate and personalized regularization to the RL reward value. Below we evaluate the design parameters $\lambda_{percent}$ and $\zeta$ as explained in \sysname.

\begin{figure}[!t]
\begin{minipage}[b]{0.65\textwidth}
\centering
\includegraphics[scale=0.48]{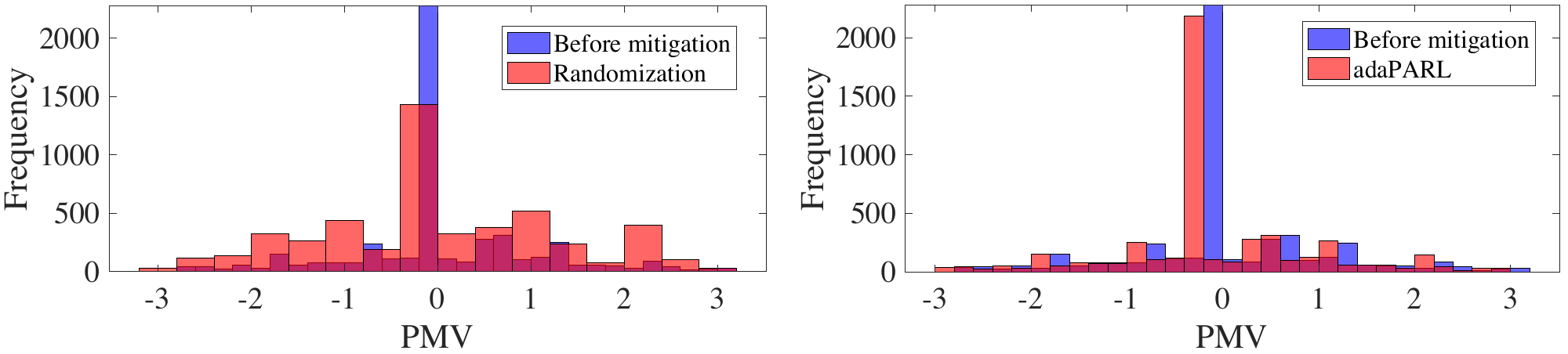}\\
\caption{$PMV$ histogram before and after mitigation. Left: randomization mitigation with $p=0.5$. Right: \sysname mitigation with $\lambda_{percent}=80\%$.}
\label{fig:pmv_bfr_aftr_rand}
\end{minipage}\hspace{2mm}
\begin{minipage}[b]{0.3\textwidth}
\centering
\includegraphics[scale=0.7]{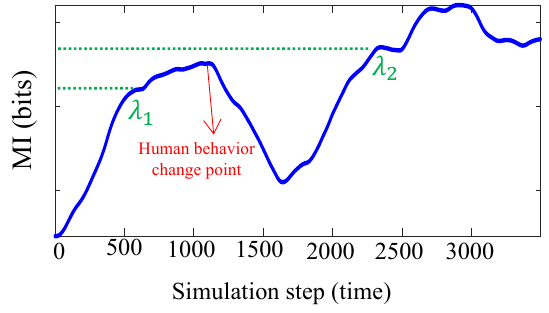}
\caption{Human changes the behavior which causes the change in MI curves. }
\label{fig:changeinbeh}
\end{minipage}
\end{figure}

\begin{itemize}[noitemsep, leftmargin=*, topsep=0pt]
\item \textbf{Tuning $\boldsymbol{\lambda_{percent}}$:} Using the same metrics we explained in Section~\ref{sec:randomization} to evaluate privacy-utility trade-off, we evaluated the effect of different values of the design parameter $\lambda_{percent}$ for the three humans as shown in Figure~\ref{fig:allbigexp1}~(row 1) with $\zeta$ at $0.6$. We observe that $\lambda_{percent}=80\%$ shows a good compromise between privacy mitigation and utility. 
Regarding privacy mitigation,  Figure~\ref{fig:cluster2}~(Bottom) presents the effect of \sysname with $\lambda_{percent}=80\%$ on clustering results compared with the randomization approach. As Figure~\ref{fig:cluster2}~(Bottom) demonstrates, the eavesdropper is less likely to predict human activity after mitigation. The clustering accuracy drops to $50\%$ for the human subject with regular activity ($H_2$) with $\zeta = 0.6$.

As for the application utility, Figure~\ref{fig:pmv_bfr_aftr_rand}~(right) shows the histogram of the total $PMV$ before mitigation and after using \sysname for the human with the regular activity schedule ($H_2$).
After using \sysname with $\lambda_{percent}=80\%$, the $PMV$ histogram around zero (best thermal comfort) does not change dramatically, and also the histogram values outside of the comfort zone decrease (in comparison with randomization), which indicates that while the privacy of the human subject is preserved, the thermal comfort of the subject is affected. 

 \item \textbf{Tuning $\boldsymbol{\zeta}$:} Similarly, as Figure~\ref{fig:allbigexp1}~(row 4) illustrates, by increasing $\zeta$, the utility (the STD of $PMV$) increases while clustering accuracy drops on average across the three humans to $50\%$ before the STD of the $PMV$ exceeds $1$. Hence, $\zeta=0.6$ provides a good compromise for the privacy-utility trade-off.

\item  \textbf{Tracking intra-human variability:} As human behavior may change over time, the value of $\lambda$ has to adapt and can not be fixed even for a single human.  
Figure~\ref{fig:changeinbeh} shows the MI for a human with a routine behavioral life pattern with a growing MI at the beginning. Human behavioral pattern changes to follow another growing MI pattern. In particular, to simulate this change in behavior,  we switched between the behavior of $H_2$ and $H_1$ at runtime. As seen in the figure, $\lambda$ follows a newly fitted MI curve to follow these changes in the human behavioral pattern. 

\end{itemize}


\subsubsection{\textbf{Mitigation 3: Comparison with the baseline}}
We compared \sysname with a baseline method proposed in ~\cite{cundy2020privacy} when the reward value is regularized (penalized) by the MI from the beginning, regardless of the shape of the MI (human behavior). In our setup, this means $\lambda_{percent}=0$. We call this approach a fixed privacy approach. Figure~\ref{fig:allbigexp1}~(row 5) provides this comparison across different values of $\zeta$. This early penalization hinders the RL agent from learning the human comfort zone partially, and the STD of the $PMV$ grows faster than \sysname. In \sysname, by tuning the parameter $\lambda$, RL has time to learn the human comfort zone, and the human experiences less uncomfortable duration (with slower $PMV$ STD growth). This pattern is followed for all three humans with different behavioral patterns, which means that \sysname can adapt to human variability with less loss in the system utility compared to the baseline method.

\vspace{-3mm}
\subsection{Observations}
In this section, we summarize the observations from application 1.  
Using \sysname with personalized $\lambda$, we could achieve privacy mitigation close to the baseline method with a smaller loss on the application utility per human. 
In particular, \sysname is able to enhance the utility by $46\%$ (STD of PMV) compared to randomization and by $57\%$ compared to the baseline method on average across three different human behavior. In terms of privacy leak mitigation, \sysname reduces the privacy leak by $16\%$ (the clustering accuracy) compared to the baseline method on average. Furthermore, compared with the randomization, the privacy leak in \sysname is decreased by $38\%$. Moreover, by using the parameters $\lambda$ and $\zeta$, \sysname can adapt to intra- and inter-human variability and regulate the privacy utility trade-off. Results showed that when privacy leak is highly mitigated (high values of $\zeta$, with $\zeta = 0.8$), the utility is in the acceptable range ($-0.5\leq PMV \leq0.5$). Hence, \sysname was able to  achieve an acceptable performance even with high privacy requirements.

%% file: 06_smartlearning.tex
\vspace{-1mm}
\section{Application 2: Human-in-the-Loop Virtual Reality Smart Classroom}\label{sec:app2}\vspace{-1mm}
The first experiment provided insights into the effect of different design parameters in \sysname in a controlled simulated environment. Next, we design a real-world VR application to evaluate \sysname.
Inspired by the recent paradigm shift in the education system post-COVID-19 era and the need for personalized and remote education setup, we selected a smart classroom IoT application using remote instruction with VR. 
During elongated training/education periods, especially in an online or remote environment, human performance is prone to significantly decline~\cite{terai2020detecting} due to distractions, drowsiness, and fatigue. In this experiment, an RL agent monitors these changes in the human state and provides personalized feedback to improve human learning performance. 
The eavesdropper is located in the cloud and has access to the actions taken by the RL model. As motivated in Experiment~1 (Section~\ref{sec:app1}), these adaptation actions are correlated to the human private state (learning performance and mental state).

We first design the application and then show the correlation between the RL agent taken actions and the human mental state. Lastly, we apply mitigation techniques, including \sysname, to mitigate the private data leak.

\vspace{-2mm}
\subsection{System Design and Implementation with VR} 
We incorporated $2$ presentation modes ($2$D and $3$D) to present the lecture contents to the participants. We used the Virtual Reality (VR) technologies for the $3$D presentation mode because recent studies showed that these new technologies would have a significant impact on the learning~\cite{ibanez2014experimenting}, and workforce training 
sectors. Using a VR device (Oculus device), we provided the $3$D presentation, and with the regular laptop screen, we provided the $2$D visualization mode. We chose the lecture contents from Khan Academy 
along with their quizzes that cover topics on biology~\cite{KhanBiology}, chemistry~\cite{KhanChemistry}, and physics~\cite{KhanPhysics}. We asked $15$ participants, all within the age range of $20-30$, to watch these lectures. 
Each lecture is stand-alone and does not require any prior knowledge from the participants to be understood. The participants' main task was to watch the lecture and pay attention to answer the questions regarding the content at the end of the lecture. We provided the $3$D version of these lectures by converting them from $2$D to $3$D for VR presentation mode. Each lecture is $\sim55$ minutes, in narrative style, and does not include any quizzes or other interruptions in the middle of them. 

\begin{figure}
\centering
\includegraphics[scale = 0.7]{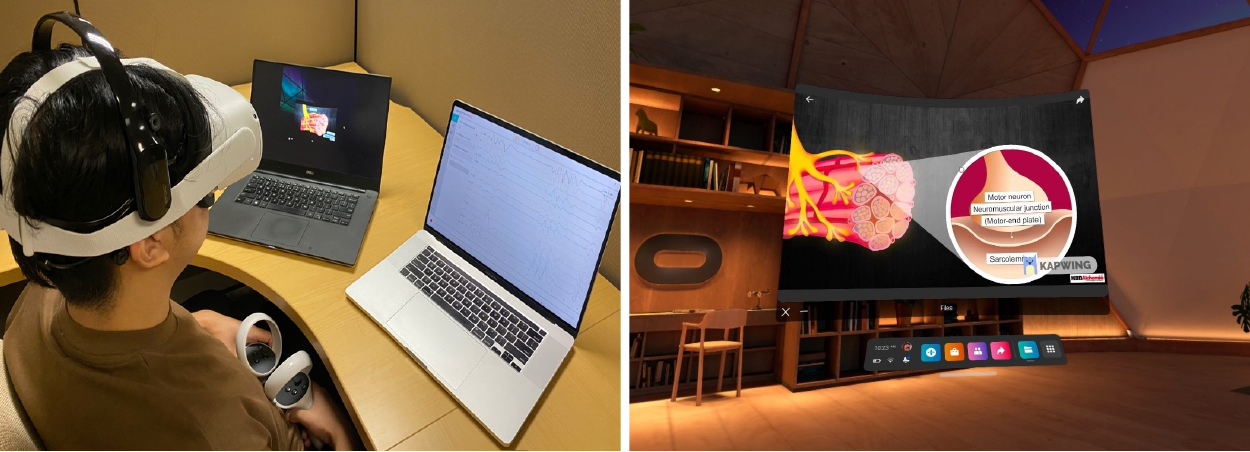}
\caption{Left: EMOTIV Epoc\_x and Oculus device worn by a participant. The laptop on the left casts the participant's view on the Oculus device, and the laptop on the right presents the live EEG signal collected from the EMOTIV device. Right: Screenshot of the view of the participant while watching biology content on an Oculus device with an office background.}
\label{fig:epoc}
\end{figure}

We used EEG wearable devices to monitor the human EEG signals to infer the mental state related to learning. We used an EMOTIV EPOC\_X 14 channels portable EEG device. Before the experiment, we presented a $10$ minutes $2$D video presentation on a laptop screen to measure their baseline mental state signals. We use every $10$ minute duration of the EEG raw data to infer the human state regarding their alertness and readiness to learn. While several wearable devices measure various physiological signals that can be used to infer the human alert level or drowsiness level, we choose the EEG signal due to the recent studies that showed the frontal lobe activation of the brain could be used to infer the human ability to learn and cognitive performance~\cite{eslinger1985severe}. 
We divided each lecture into $10$ minutes videos that we call stages, with approximately $5$ stages each lecture\footnote{The average attention span of the human is $10$ to $15$ minutes.}. Figure~\ref{fig:epoc} shows the setup of the application.

\vspace{-2mm}
\subsection{RL Design}
\subsubsection{\textbf{Human State Space}}\label{sec:dizzy} 
Albeit the beneficial aspects of VR technologies in education and workplace training, humans react differently to the VR environment. Some humans report vertigo and cybersickness symptoms during exposure to VR~\cite{brunnstrom2020latency}, which can affect the learning experience. 
We designed the human state as a combination of three features, including alertness level (AL), fatigue level (FL), and vertigo level (VL). Accordingly, we classify these three features into binary classes to reduce our state space. In particular, \emph{AL} is classified as ``Alert (1)'' versus ``Not Alert (0)'', \emph{FL} is classified as ``Fatigue (1)''  versus ``Vigor (0)'', and \emph{VL} is classified as ``Not Vertigo (1)'' versus ``Hypocalcemia (0)''. $\mathcal{S}$ refers to a tuple of $3$ features: $\mathcal{S} = \{(AL,FL,VL) : AL \in \{0,1\},\ FL \in \{0,1\},\ VL \in \{0,1\} \}$. Table~\ref{tab:state} illustrates the state space. Each feature is classified based on a corresponding threshold $ \delta_{AL}, \ \delta_{FL}, and \ \delta_{VL}$. If the feature's measured value passes the threshold, the class for the given feature is $1$; otherwise, it is $0$. Accordingly, the best human state for learning is $S_8$, where the human is alert and is not experiencing cybersickness (no fatigue and vertigo). In contrast, the worst state is $S_1$, where the human is not alert and experiencing cybersickness (fatigue and vertigo).

Indeed, humans can transition between any of these states. It is worth mentioning that these thresholds can be tuned based on the application and participant. Below we describe how we infer the human state and calculate these thresholds.

\begin{table}[!t]
\caption{Human state is one of 8 states depending on the alertness level (AL), fatigue level (FL), and vertigo level (VL). } 
\label{tab:state}
\begin{tabular}{
>{\columncolor[HTML]{cbcfd1}}c 
>{\columncolor[HTML]{cbcfd1}}c 
>{\columncolor[HTML]{cbcfd1}}c 
>{\columncolor[HTML]{cbcfd1}}c 
>{\columncolor[HTML]{cbcfd1}}c 
>{\columncolor[HTML]{cbcfd1}}c 
>{\columncolor[HTML]{cbcfd1}}c 
>{\columncolor[HTML]{cbcfd1}}c 
>{\columncolor[HTML]{cbcfd1}}c}
\textbf{AL}  & \cellcolor[HTML]{f0f5f5}1 & \cellcolor[HTML]{f0f5f5}1  & \cellcolor[HTML]{f0f5f5}1  & \cellcolor[HTML]{f0f5f5}1   & \cellcolor[HTML]{f5cbcb}0 & \cellcolor[HTML]{f5cbcb}0 & \cellcolor[HTML]{f5cbcb}0  & \cellcolor[HTML]{f5cbcb}0  \\

\textbf{FL} & \cellcolor[HTML]{f0f5f5}1 & \cellcolor[HTML]{f0f5f5}1 & \cellcolor[HTML]{f5cbcb}0 & \cellcolor[HTML]{f5cbcb}0 & \cellcolor[HTML]{f0f5f5}1 & \cellcolor[HTML]{f0f5f5}1 & \cellcolor[HTML]{f5cbcb}0 & \cellcolor[HTML]{f5cbcb}0  \\

\textbf{VL} & \cellcolor[HTML]{f0f5f5}1 & \cellcolor[HTML]{f5cbcb}0 & \cellcolor[HTML]{f0f5f5}1 & \cellcolor[HTML]{f5cbcb}0 & \cellcolor[HTML]{f0f5f5}1 & \cellcolor[HTML]{f5cbcb}0 & \cellcolor[HTML]{f0f5f5}1 & \cellcolor[HTML]{f5cbcb}0  \\

\textbf{State} & \cellcolor[HTML]{cbcfd1}\textbf{S8} & \cellcolor[HTML]{cbcfd1}\textbf{S7} & \cellcolor[HTML]{cbcfd1}\textbf{S6} & \cellcolor[HTML]{cbcfd1}\textbf{S5} & \cellcolor[HTML]{cbcfd1}\textbf{S4} & \cellcolor[HTML]{cbcfd1}\textbf{S3} & \cellcolor[HTML]{cbcfd1}\textbf{S2} & \cellcolor[HTML]{cbcfd1}\textbf{S1}
\end{tabular}
\end{table}

\begin{itemize}[topsep=0pt, noitemsep, leftmargin=*]
    \item \textbf{Alertness level (brain engagement) ($AL$):} To measure human alertness and engagement during the learning process, spectro-temporal EEG signal analysis can be used. One method recently implemented to analyze the EEG signal during the learning process is fractal dimension~\cite{foroutan1999advances}. 
    Various methods have been developed to calculate the fractal dimension, mainly based on the entropy concept. In this experiment, we used the box-counting method to calculate the fractal dimension~\cite{foroutan1999advances} on the recorded EEG time series. Since the frontal lobe of the brain is responsible for cognitive functions such as memory and problem solving~\cite{eslinger1985severe}, we used the $F3$ and $F4$ channels of the EEG device (located in the frontal lobe). We averaged the outcomes of these channels for our analysis.
    After preprocessing (filtering and noise removal) and fractal dimensions analysis, we observed that the fractal dimension of the EEG signal is higher in the $3$D compared to the $2$D videos. Since the fractal dimension reflects the complexity of the signal, this result indicates that the EEG signal is more complex in response to $3$D visual stimuli than $2$D visual stimuli. In other words, the human brain becomes more engaged with a stimulus when presented in the $3$D compared to the $2$D. We used this measurement as the $AL$ state.   
    
    \item \textbf{Fatigue Level ($FL$):}   
    We define a measure for $FL$ that exploits the EEG signals. In particular, we use Wavelet Packet Decomposition (WPD) to decompose the EEG signal into its spectral sub-bands with $1$~$Hz$ resolution. 
    Recent work in the literature showed that fatigue and drowsiness correlate with the range of $8-14$ $Hz$, which is mostly the $\alpha$ band of the EEG signal~\cite{correa2014automatic}. 
    Accordingly, we define $FL$ as the power spectrum of the $\alpha$ band. Similar to $AL$, we collected the data from $F3$ \& $F4$ channels and averaged the resulting $FL$ from both channels. 
    
    \item \textbf{Vertigo Level ($VL$):} The most commonly reported measure of vertigo and cybersickness symptoms is the Simulator Sickness Questionnaire (SSQ). The SSQ was derived directly from the Pensacola Motion Sickness Questionnaire (MSQ)~\cite{golding1998motion}. The MSQ consists of a list of $25$ to $30$ symptoms, such as spinning, vertigo, and may vomit. Symptoms severity are rated on four levels, ``none'' (0), ``slight'' (1), ``moderate'' (2), and ``severe'' (3). A total score was computed by summing item scores. The highest score was determined to specify emesis as the worst case of sickness. In particular, the SSQ is a selection of $16$-items 
    from the MSQ with a different scoring scheme.  
    Based on three main subfactors of cybersickness, including Nausea ($N$), Oculomotor ($O$), and Disorientation ($D$),
    a Total Score (TS) is computed, representing the overall severity of cybersickness experienced by the subject.  
    In particular, $TS$ can range from $0$ to $235.62$~\cite{stone2017psychometric}.  
    Participants were asked to fill out the SSQ questionnaire, and we used it to calculate the $TS$. 
    The threshold for $TS$ to consider it vertigo depends on the application~\cite{stone2017psychometric}).
    
    \item \textbf{Choosing the thresholds ($\delta_{AL}$, $\delta_{FL}$, $\delta_{VL}$):} $\delta_{AL}$ was defined based on the calculation of the $AL$. $AL$ correlates with the fractal dimension values calculated for the EEG signal. We compared each calculated fractal dimension value with the baseline (measured before the experiment).
    If the $AL$ exceeds the baseline ($\delta_{AL}$), the human is classified as ``Alert'' (1), else the human is classified as ``Not Alert''(0).

    We used the spectral power of the $\alpha$ band to define the threshold for $FL$. $\delta_{FL}$ is the midpoint of the $t$ measure, calculated based on the $p-Values$ of the comparison of the $\alpha$ band of the baseline EEG and the current stage EEG signals. If the $t$ measure exceeds $\delta_{FL}$, then the human is classified as ``Fatigue'' (0), or else the human is classified as ``Not Fatigue''(1).

    We defined $\delta_{VL}$ to be $\frac{TS_{max}}{4}$ where $TS_{max}=235.62$. If the measured $TS$ is bigger than the $\delta_{VL}$, the human is classified as ``Not Vertigo'' (0), or else the human is classified as ``Vertigo''(1). At the end of each stage of the experiment, participants are asked to fill out the SSQ, and using the questionnaire, we calculate $TS$. 

\end{itemize}

\vspace{-2mm}
\subsubsection{\textbf{Action Space}}\label{sec:policy}
The action space $\mathcal{A}$ in this VR application includes the following actions: (1): Give a break to the human, (2): Enable VR mode by switching from $2$D to $3$D, (3): Disable VR mode by switching from $3$D to $2$D, (4): Changing the content of the presentation, and (5): No change to the learning environment.

In particular, enabling the VR mode increases brain engagement and enhances learning performance. However, some humans may experience cybersickness with exposure to VR; hence the RL agent may need to switch back to regular $2$D to reduce cybersickness symptoms. Moreover, a break during a learning session may also be needed to reduce drowsiness, cybersickness symptoms, or cognitive load. Hence, $\mathcal{A}$ is bounded and discrete and can be within the range $[1,5]$:
$\mathcal{A} = \{a : a \in [1,5], a \in \mathbb{N}\}.$

\vspace{-2mm}
\subsubsection{\textbf{Reward}} We used the same definition of the reward function explained in Section~\ref{sec:adaparl}. The human performance in a quiz dictates the reward value after every learning module, where the score in this quiz is measured as a percentage. Quiz quantification applies to $10$ multiple-choice questions uniformly as the $10/10 \ (100\%)$ and $0/10 \ (0\%)$ scores receive rewards $100$ and $0$, respectively. 

\vspace{-3mm}
\subsection{Private Information Leak} 
As described in our threat model in Section~\ref{sec:threatmodel}, eavesdroppers in the cloud can access the RL desired actions and run machine learning algorithms to infer the participant's private information. The human state is calculated at the edge. The human mental state is private information, and any gain by the attacker to this information is considered a privacy intrusion. 
Similar to the analysis we did in the first application in Section~\ref{sec:app1}, we monitor the MI between the RL action and the participants' state. Figure~\ref{fig:misecapp2} 
illustrates the MI between taken actions and states of a participant (solid red line referring to ``before mitigation''). The MI reaches approximately $\approx 1.5~ bits$. 
A similar approach for an eavesdropper we designed in the first application in Section~\ref{sec:privacyleak} can be deployed in this application. An eavesdropper 
can use unsupervised learning techniques, such as clustering. Hence, we will evaluate \sysname's ability to mitigate the private leak.

\vspace{-3mm}
\subsection{Privacy Leak Mitigation with \sysname}
As mentioned in the previous experiment, we aim to mitigate this information leak using two approaches, action randomization, and \sysname. 

\vspace{-2mm}
\subsubsection{\textbf{Mitigation 1: Randomization}}
\begin{figure}[!t]
\raisebox{5 ex}{\begin{minipage}[b]{0.3\textwidth}
\centering
\includegraphics[scale=0.43]{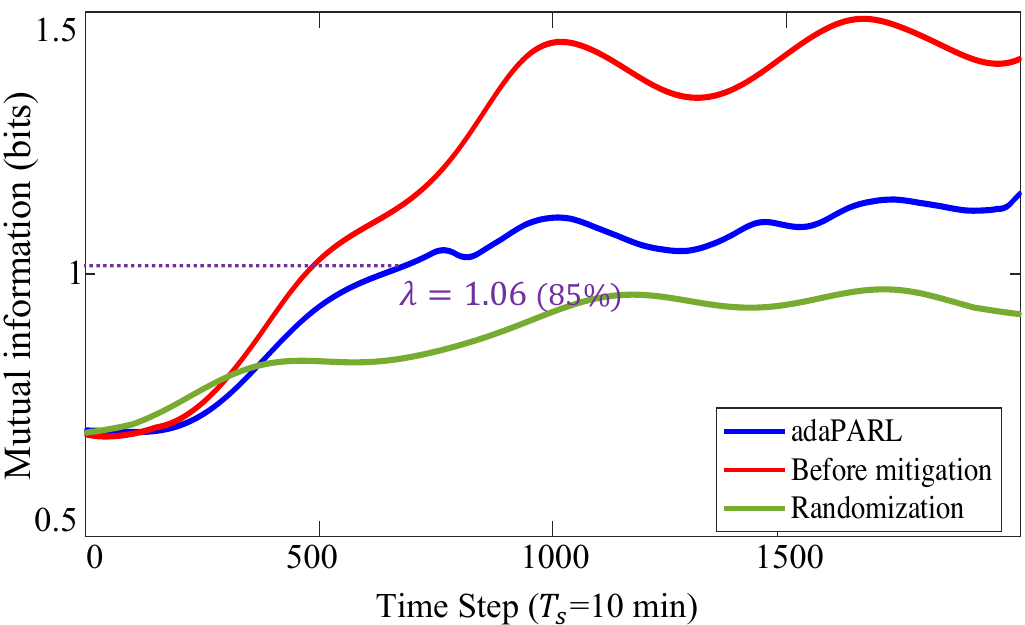}
\caption{MI between actions and participant's states before and after adding mitigation using randomization and \sysname at $\lambda =1.06 \ (85\%)$.}
\label{fig:misecapp2}
\end{minipage}}\hspace{2mm}
\begin{minipage}[b]{0.3\textwidth}
\centering
\includegraphics[scale=0.42]{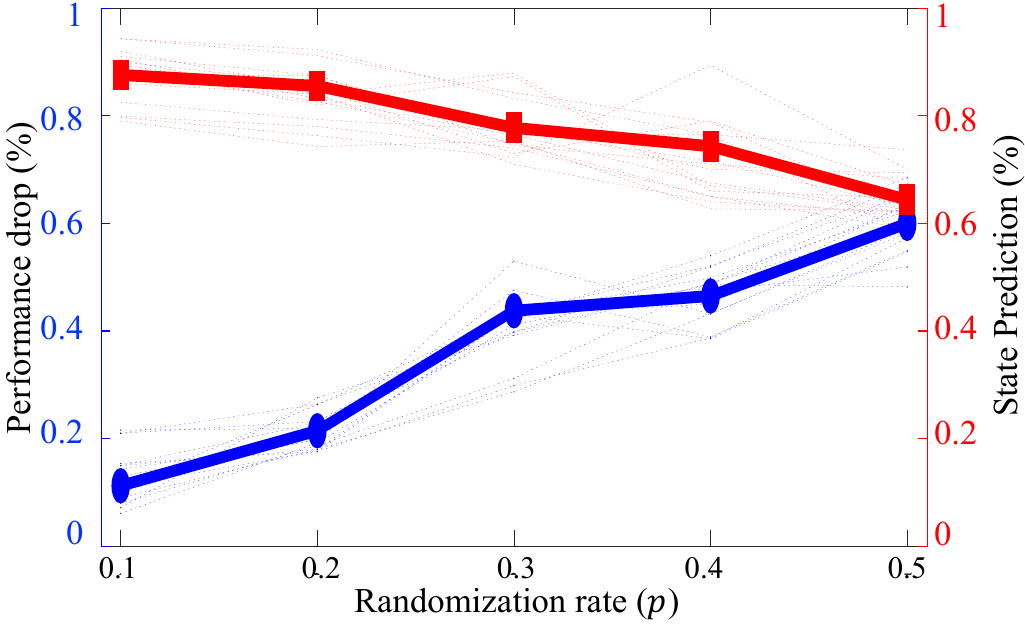}\\
\caption{Trade-off between state prediction accuracy (\%) and performance drop on the quiz (\%) after adding randomization. The dotted line presents one participant, and the solid lines present the averages for $15$ participants.}
\label{fig:exp2results1}
\end{minipage}\hspace{2mm}
\raisebox{5 ex}{\begin{minipage}[b]{0.3\textwidth}
\centering
\includegraphics[scale=0.44]{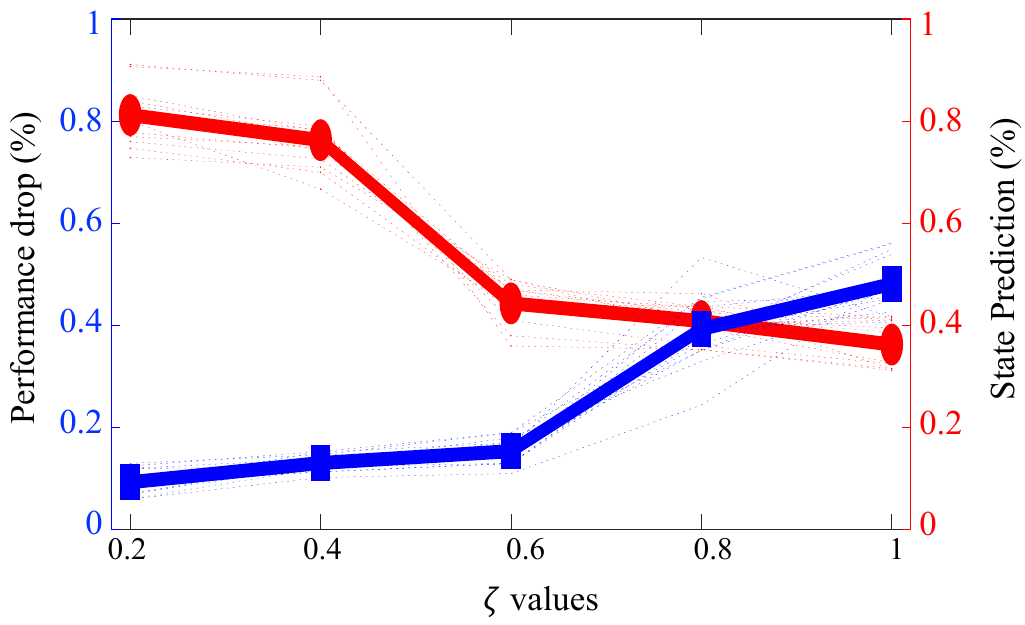}\\
\caption{Trade-off between participants' state prediction accuracy (\%) and performance drop on the quiz (\%) using \sysname ($\lambda_{percent} = 85\%$).}
\label{fig:exp2results2}
\end{minipage}}
\end{figure}

As a representative result, by choosing random actions ($p=0.5$), Figure~\ref{fig:misecapp2}  
depicts the randomization effect on MI between RL actions and one of the participant's states, which shows the decrease in MI from $\approx 1.5~bits$ to less than $\approx 1~bits$.  
We evaluate the privacy-utility trade-off using action randomization. In particular, we use the drop in the performance in the quizzes as a utility metric. We use the human state prediction accuracy using clustering (similar to application 1) as the privacy leak measure. We emphasize here that even though the eavesdropper may not know the actual human state, the change in the human state (learning pattern) through observed actions can be inferred, which can leak private information, such as the human attention span. 

To study the effect of randomization $p$ without asking the participant to repeat the experiment $5$ times, which may bias the results, we used the data we collected from the online experiment with the 15 participants at $p=0.5$. In particular, from the online experiment, we knew the quiz performance of a participant given a state-action pair. Hence, we change the value of $p$ offline to generate different actions per state and record the expected quiz performance. Figure \ref{fig:exp2results1}  shows the effect of parameter $p$ on utility-privacy trade-off for the randomization algorithm for the participants (dotted lines) and the average (solid line), which illustrates a reduction in the state prediction accuracy on average to $60\%$ with a $60\%$ drop in utility (performance).
Adding more random actions leads to lower performance in the quiz as the chosen presentation mode is less likely to be optimal for the participant's current state. This trade-off shows how privacy protection increases according to the randomization of the actions. It also demonstrates that performance drops quickly after increasing randomization.

\begin{table}
\centering
\caption{Privacy (State Prediction) vs. Utility (Performance Drop) trade-off and $\lambda$ parameter results of \sysname for all the $15$ participants.}
\begin{adjustbox}{width=\columnwidth,center}
\label{app2param}
\small
\begin{tabular}{lcccccccccccccccc}
\hline
\textbf{Parameter}     & \textbf{P1} & \textbf{P2} & \textbf{P3}   & \textbf{P4}    & \textbf{P5} & \textbf{ P6} & \textbf{P7} & \textbf{P8} & \textbf{P9}   & \textbf{P10}    & \textbf{P11} & \textbf{ P12} & \textbf{P13} & \textbf{P14} & \textbf{P15}   & \textbf{Ave} \\\hline
\textbf{$\lambda$ (with $\lambda_{percent}=85\%$)}      & 1.06  & 1.11  &  0.98 & 0.97 & 1.03  &  0.93 & 1.02  & 1.01  & 1.04 & 0.95 & 0.91  &  1.02 & 1.16   &  0.89 & 0.96 & \textbf{1.01}  \\
\textbf{State Prediction (\%)} & 0.41  & 0.38  & 0.46  & 0.47 &  0.43 &  0.48 &  0.43 &  0.44 & 0.43 & 0.47  & 0.49 &  0.44 & 0.36   & 0.49  &  0.47 & \textbf{0.44} \\
\textbf{Performance Drop (\%)} & 0.15  & 0.13  & 0.18  & 0.17 & 0.15  & 0.16  & 0.14  &  0.16  & 0.15 & 0.17 & 0.19  & 0.13 & 0.11   & 0.19   & 0.13  & \textbf{0.15} \\\hline
\end{tabular}
\end{adjustbox}
\end{table}

\vspace{-2mm}
\subsubsection{\textbf{Mitigation 2: \sysname}}
We tuned the parameter $\lambda_{percent}$ on $5$ participants offline using a similar approach to application 1. We used the same category of lecture content (i.e., biology) but not the same content as the one we used in the online experiment to prevent any content bias on the participants. The offline test resulted in $\lambda_{percent}=85\%$. Hence, we set $\lambda_{percent}=85\%$ across the $15$ participants during the online test. In Figure~\ref{fig:misecapp2} (solid blue line), we show the MI between the actions and one of the participant's states using \sysname. As Figure~\ref{fig:misecapp2}   
illustrates, the MI increases as the agent learns to take corrective actions and then using $\lambda = 1.06$  ($\lambda_{percent} =85\% $) 
\sysname regularized the growth of the MI as explained in Section~\ref{sec:adaprivacy}. 

Similar to the approach we used to study the effect of $p$ in the randomization, we study the effect of $\zeta$. 
Figure~\ref{fig:exp2results2} presents the percentage of performance drop and participants' state prediction accuracy by changing the values of $\zeta$ for the participants (dotted lines) and its average (solid line).  
Increasing the parameter $\zeta$ decreases the eavesdropper's ability to predict the participants' state. The prediction ability of the eavesdropper for $\zeta~=~ 0.6$ decreases by $\sim 50\%$. At the same time, the participant's performance is higher than $75\%$  (performance drop is less than $25\%$), showing improvement in the trade-off compared to the randomized actions.       
Table \ref{app2param} provides the numeric details of the trade-off parameter ($\zeta$) and $\lambda \ (\lambda_{percent}=85\%)$ values specific to each participant. \sysname chooses different values of $\lambda$ for each participant (inter-human variability). 

\vspace{-3mm}
\subsection{Observations}
Results from this application show that using the parameter $\zeta$ \sysname can mitigate the eavesdropper's ability to predict the participants' state. Our results show that the prediction ability of the eavesdropper for $\zeta~=~ 0.6$ decreases by $\sim 50\%$. On the other hand, the utility (participant's performance in a quiz) is higher than $75\%$, meaning that \sysname can improve the trade-off compared to using the randomization approach.       
Furthermore, \sysname uses different numeric values of $\lambda$ for different participants, which shows that \sysname can provide adaption to inter-human variability.
On average, the accuracy of the private state detection decreases to $44\%$ before the drop in participants' performance (utility) passes $15\%$. In contrast, in the randomization approach, the $20\%$ drop in the utility leads to a high privacy cost of $80\%$ as indicated in the prediction accuracy. In randomization, the privacy leak, which is the prediction accuracy, never exceeds $65\%$, even with the degradation in the utility. For instance, when the participants' performance drops by over $60\%$, the privacy leak is still $65\%$.  

%% file: 08_conclusion.tex
\vspace{-2.5mm}
\section{Conclusion}\vspace{-1mm}
In this paper, we proposed \sysname, an adaptive human-in-the-loop privacy-aware RL algorithm that addresses the privacy challenges associated with human variability in RL-based systems trained with privacy-sensitive data in IoT applications. We adopted a typical edge-cloud threat model architecture where all the sensitive human state inference is calculated on a trusted edge that hosts the RL agent, and the cloud has only access to the desired control actions where an eavesdropper is mounted.  
\sysname provided an adaptive and personalized threshold ($\lambda$) to regularize the reward function of the RL agent, which changes its value at runtime based on the changes in human behavior to mitigate the privacy leak. 
We validated \sysname on two Human-in-the-Loop IoT applications in simulated (smart house) and real-world (VR smart classroom) environments. We showed that \sysname could achieve a personalized privacy-utility trade-off through two tunable design parameters, $\zeta$, which provides the privacy-utility trade-off, and $\lambda$, which provides the adaptation to inter-human variability.  
In the first application, on average, \sysname improved the application utility over the randomization approach by $43\%$ and over the baseline approach by $57\%$. 
Furthermore, \sysname reduced the privacy leak on average by $23\%$. We implemented \sysname in a real-world application and demonstrated how \sysname is capable of adapting to inter-human variability. Thanks to the flexibility of its parameter ($\lambda$), \sysname was able to adapt to $15$ different human participants. Furthermore, \sysname provided a tunable design parameter ($\zeta$) to provide the flexibility to choose the desired privacy-utility trade-off per application.